\documentclass[11pt]{article}

\usepackage[final]{acl}

\usepackage{times}
\usepackage{latexsym}
\usepackage{listings}  
\usepackage[T1]{fontenc}
\usepackage{ragged2e}
\usepackage{graphicx}    
\usepackage[utf8]{inputenc}

\usepackage{microtype}
\lstdefinestyle{promptstyle}{
    basicstyle=\small\ttfamily,
    breaklines=true,
    frame=single,
    captionpos=b,
    keepspaces=true,
    numbers=none,
    showspaces=false,
    showstringspaces=false,
    showtabs=false,
    tabsize=2,
    backgroundcolor=\color{gray!10},
    columns=flexible,
    breakatwhitespace=true,
    postbreak=\mbox{\textcolor{red}{$\hookrightarrow$}\space},
}
\usepackage{inconsolata}
\usepackage{inconsolata}
\usepackage{subcaption}
\usepackage{graphicx}
\usepackage{booktabs}
\usepackage{amssymb}
\usepackage{makecell}
\usepackage{xspace,mfirstuc,tabulary}
\usepackage{tabularray}
\UseTblrLibrary{booktabs}
\usepackage{pifont}
\usepackage{amsmath}
\usepackage{multirow}
\usepackage{enumitem}
\usepackage{tabularx} 
\usepackage{array}    
\usepackage[table]{xcolor}
\usepackage{xcolor}
\usepackage{soul}
\usepackage{natbib}

\newcommand{\projectname}{\texttt{CollabSim}\xspace}
\title{\projectname: A CSCW-Grounded Methodology for Investigating Collaborative Competence of LLM Agents through Controlled \\ Multi-Agent Experiments}

\author{
Jiaju Chen \\
Northeastern University
\And 
Bo Sun \\
Northeastern University
\And 
Yuxuan Lu \\
Northeastern University
\AND
Yun Wang \\
Microsoft Research Asia
\And 
Dakuo Wang \\
Northeastern University 
\And 
Bingsheng Yao\thanks{~Corresponding Author: \href{mailto:b.yao@northeastern.edu}{b.yao@northeastern.edu}. } \\
Northeastern University
}

\begin{document}
\maketitle
\begin{abstract}
Multi-agent systems (MAS) built on large language models have shown growing promise, with their effectiveness resting on agents' ability to coordinate through text-based channels much as human teams do. Yet recent study suggests that MAS often falter not because agents lack individual task-solving ability, but because they lack collaborative competence: the capacity to establish common ground, maintain shared task understanding, balance individual and collective incentives, and repair misalignment as interaction unfolds. 
Decades of research in Computer-Supported Cooperative Work have characterized these requirements for human teams coordinating under constrained communication, yet existing MAS evaluations focus mainly on task outcomes or single-agent proficiency in reasoning, planning, and tool use. 
To enable a systematic analysis of agents' collaborative competence in MAS, we introduce \projectname, a configurable simulation framework that combines a theory-grounded definition of collaborative capabilities, controlled manipulation of interaction conditions, and action-level probing of agents' internal states.
Experiments across four LLMs show that \projectname can capture condition effects, separate model performance patterns, and reveal task-dependent effects of agent design.
\end{abstract}

\section{Introduction}

\begin{figure}[th!]
    \centering
    \includegraphics[width=0.99\linewidth]{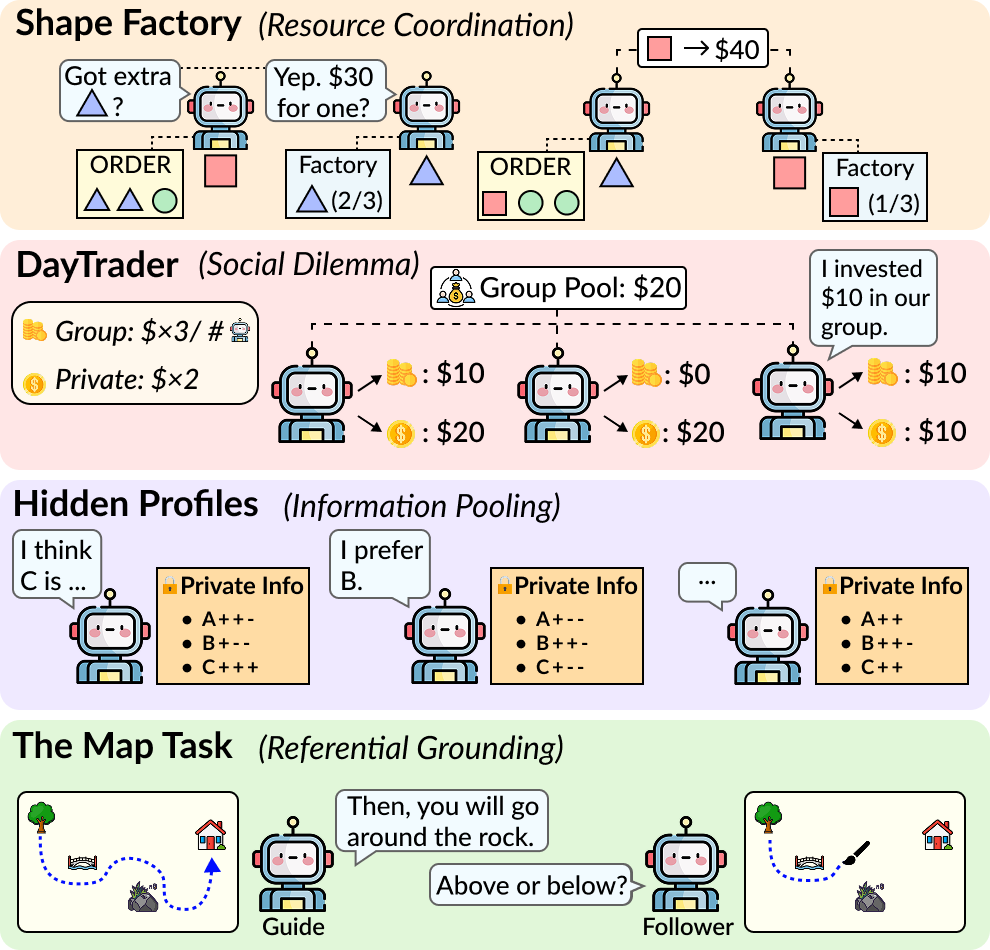}
    \caption{Illustrations of the four multi-agent experiments instantiated in \projectname: Shape Factory~\cite{bos2004group},  DayTrader~\cite{bos2002effects}, Hidden Profile~\cite{stasser1985pooling}, and The Map Task~\cite{anderson1991hcrc}.}
    \label{fig:experiments}
    \vspace{-4mm}
\end{figure}

A growing body of work has explored multi-agent systems (MAS) for complex task solving, with applications in software engineering \cite{he2025llm}, scientific research \cite{gottweis2025towards}, and web automation \cite{Zhou2023WebArenaAR}. 
As these systems tackle increasingly complex tasks, their effectiveness increasingly rests on the collaborative mechanism among agents, where multiple agents with distinct roles perform individual tasks, exchange information through text-based channels, and coordinate toward shared goals \cite{chen2025multi, Zou2025LLMBasedHC}. 
The ability of individual agents to collaborate effectively with their partners thus becomes as consequential as their ability to solve tasks independently \cite{wang2024n}.

A parallel line of evidence, however, suggests that MAS often fail not because individual agents lack task-solving capacity, but because they struggle to coordinate during collaboration \cite{cemri2026multi, li2026flowsteer}.
Agents have been shown to misread partners' states and intentions \cite{mu2026adaptive}, lose track of shared task progress as context accumulates \cite{pappu2026multi}, fail to repair misalignment \cite{yadav2026more}, and struggle to balance individual and collective incentives \cite{tewolde2026coopeval}. 
We refer to this broad capacity as agents' \textbf{collaborative competence}.

However, existing evaluation methodologies are not designed to diagnose process-level failures in agents' collaborative competence. 
Current MAS benchmarks primarily measure outcome-centric task-solving performance \cite{liu2024agentbench, zhu-etal-2025-multiagentbench, sun2025collab} or assess individual agents' proficiency in reasoning, planning, and tool use \cite{qin2023toolllm, yao2024tau}. 
Yet, these evaluations are insufficient for diagnosing the coordination breakdowns identified above, since those failures arise from agents interaction process rather than from any single agents' capability deficit.

Notably, the coordination failures described above closely parallel long-standing problems studied in Computer-Supported Cooperative Work (CSCW) research on distributed human teams, where collaborators similarly coordinate through structured text-based channels without co-presence. 
Research in this tradition has identified the core competencies required for effective remote collaboration, including common ground maintenance \cite{clark1991grounding}, shared task understanding \cite{cannon1993shared}, and misalignment repair \cite{gutwin2002descriptive}.
Prior work has also shown that collaborative competencies vary with interaction conditions like communication bandwidth, information visibility, and team structure \cite{bos2002effects, daft1986organizational}. 
To assess these competencies under controlled conditions, CSCW researchers developed experimental paradigms that isolate specific dimensions of collaborative competence \cite{bos2002effects, bos2004group, stasser1985pooling, anderson1991hcrc}, thereby providing a validated methodology for evaluating the same competencies in LLM agents.

Drawing on this methodology, we introduce \projectname\footnote{Code is available at \url{https://github.com/neuhai/CollabSim}.}, a configurable simulation framework for systematically assessing agents' collaborative competence under controlled interaction conditions. 
\projectname allows researchers to vary task constraints such as communication bandwidth, information visibility, and team size, so that the effects of specific interaction conditions on collaborative behavior can be examined. 
It also includes a probing module that elicits each agent's reported mental model awareness of the task state, partner intentions, and own reasoning after every action, enabling analysis of agents' internal collaborative states beyond observable behavior.
To demonstrate the framework's coverage, we implement four CSCW-inspired collaborative tasks that reflect the process-level challenges discussed above: resource coordination under cost asymmetry (Shape Factory; \citealt{bos2004group}), social-dilemma negotiation between individual and collective incentives (DayTrader; \citealt{bos2002effects}), distributed information pooling under information asymmetry (Hidden Profile; \citealt{stasser1985pooling}), and referential grounding through language-only spatial communication (Map Task; \citealt{anderson1991hcrc}). 

We validate \projectname across the four tasks under systematically varied interaction conditions, comparing two agent designs (persona-based and Collaboration-Theory-Informed) across four model backbones (Qwen3.6-35B-A3B, Llama-4-Maverick-17B-128E-Instruct-FP8, GPT-5.5, and Claude 4.6 Sonnet). 
The results suggest that \projectname provides stable and interpretable measurements: condition manipulations shift collaboration metrics in expected directions across models, the same task settings separate proprietary from open-source models, and cross-task evaluation reveals task-dependent effects of agent design.

\section{Related Work}

\subsection{LLM-Based Multi-Agent Systems}

Early LLM-based multi-agent systems (LLM-MAS) \cite{wu2024autogen, qian2024chatdev, hong2024metagpt} often instantiate agents with predefined personas and communication protocols. 
More recent work \cite{chen2024agentverse, chen2025multi, zhang2025aflow} has explored dynamic multi-agent architectures, where task decomposition, role assignment \cite{wang2025megaagent}, agent dependencies, and execution workflows \cite{yang2026agentnet} can be adapted based on user input. 
These systems have been applied across various domains~\cite{fang2026mlzero, zhou2026reagent}, including software engineering \cite{he2025llm}, web automation \cite{workarena}, scientific reasoning \cite{gottweis2025towards}, etc.
These LLM-MAS primarily treat agent-agent collaboration as a design strategy for improving task completion \cite{tran2025multi, cemri2026multi} instead of assessing or improving agents' collaborativeness for effective coordination.  

\subsection{Frameworks for Agent Collaboration}

Recent work has developed a variety of frameworks for evaluating and simulating multi-agent collaboration. 
The first line introduces task-based evaluation benchmarks. 
These frameworks place multiple LLM agents in collaborative settings and assess their performance across different scenarios and interaction protocols \cite{yu-etal-2025-table, sun2025collab, orogat2026understanding, tewolde2026coopeval}. 
MultiAgentBench \cite{zhu-etal-2025-multiagentbench}, for example, covers a range of cooperative tasks and measures team-level outcomes under varied coordination structures. 
Another line of work focuses on role-play-based social simulation environments, where LLM agents are instantiated with personas, social goals, and relationship contexts to simulate dyadic \cite{wang2024sotopia, wang2026mascot} or population-level interactions \cite{park2024generative, yang2024oasis, Piao2025AgentSocietyLS}. 
For instance, SOTOPIA \cite{zhou2024sotopia} provides an open-ended social interaction environment for evaluating agents’ social intelligence. 

These frameworks demonstrate that LLM agents can coordinate, communicate, and behave socially in structured settings. 
However, existing benchmark frameworks \cite{wang2024battleagentbench, sun2025collab} primarily evaluate agents' task-solving outcomes and overlook the effectiveness of agent collaboration and the coordination among agents. 
Social simulation frameworks \cite{zhou2024sotopia, Piao2025AgentSocietyLS} expose rich agent behavioral traces, but they often treat collaboration conditions as fixed design choices rather than experimental variables.

\begin{figure}[t!]
    \centering
    \includegraphics[width=0.99\linewidth]{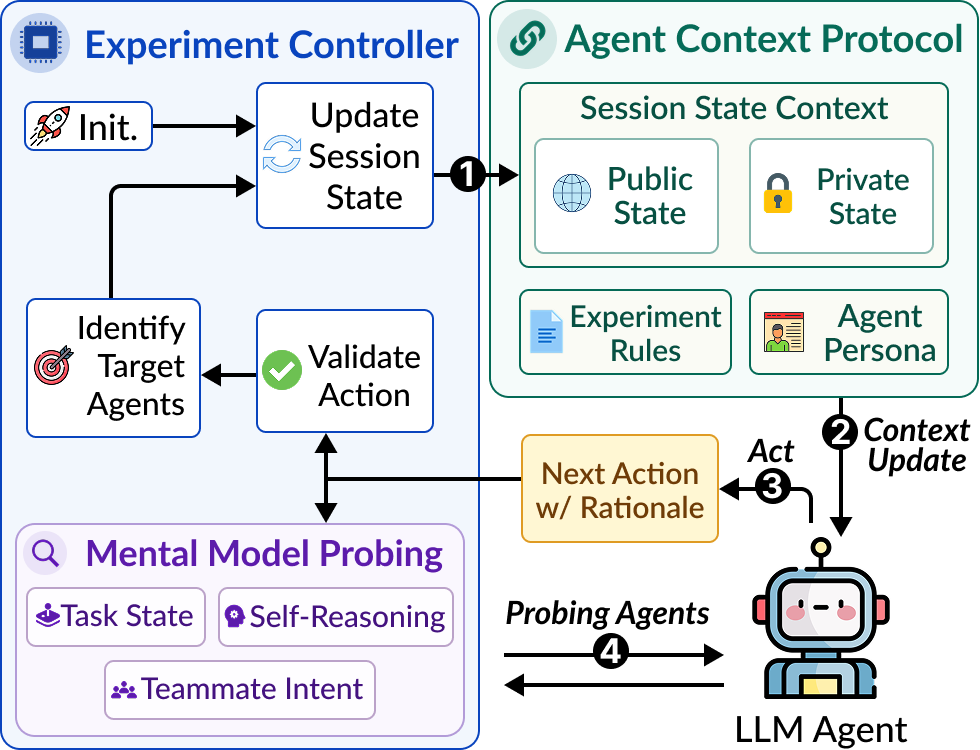}
    \caption{Architecture of \projectname. The Controller manages the action validation, state update, and mental model probing for each agent.}
    \label{fig:modules}
\end{figure}

\section{\projectname}
\label{sec:framework}

We present \projectname, a theory-grounded simulation framework for systematic investigation of multi-agent collaborative competence across classic CSCW experimental paradigms.  
We describe the system architecture that governs how agents perceive, act, and are probed (Sec.~\ref{sec:architecture}), the configurable interaction conditions that serve as experimental variables (Sec.~\ref{sec:conditions}), the probing module that captures agents' internal states (Sec.~\ref{sec:probing}), and the four collaborative tasks that cover distinct dimensions of collaborative competence (Sec.~\ref{sec:tasks}).

\subsection{System Architecture}
\label{sec:architecture}

As shown in Figure \ref{fig:modules}, \projectname organizes its architecture into two layers: an \textbf{interaction layer} that governs how agents perceive the task environment, and a \textbf{control layer} that orchestrates experiment configuration, execution, and evaluation. Figure~\ref{fig:modules} illustrates how these layers interact during one complete agent action cycle.
Formally, a collaboration experiment is defined as a tuple
\[
\mathcal{E} = (\mathcal{A}, \mathcal{S}, \{\mathcal{X}_i\}_{i=1}^n, \{\mathcal{O}_i\}_{i=1}^n, T),
\]
where \(\mathcal{A}=\{a_1,\ldots,a_n\}\) is the set of agents,
\(\mathcal{S}\) is the state space,
\(\mathcal{X}_i\) is the action space available to agent \(a_i\),
\(\mathcal{O}_i\) is the observation space of agent \(a_i\),
and \(T\) is the termination condition.
At each turn \(t\), the task maintains a shared state \(s^t \in \mathcal{S}\). 
Each agent \(a_i\) receives an observation \(o_i^t \in \mathcal{O}_i\), selects an action \(x_i^t \in \mathcal{X}_i\), and the state transitions to \(s^{t+1}\) according to task-specific dynamics.


\paragraph{Agent Context Protocol.} 
The Agent Context Protocol defines the interface through which agents perceive the task environment and act. 
In the beginning, the agent context protocol is initialized with the experiment rules and agent persona.
At each turn $t$, the system constructs each agent's observation \(o_i^t\) by combining two components: a \textit{public update}, which reflects the current shared task state $s^t$ filtered by information-visibility rules, and a \textit{private state}, which carries agent-specific facts such as remaining production capacity or assigned candidate materials.
The agent then selects an action \(x_i^t\) from the action space \(\mathcal{X}_i\).
By standardizing how context is constructed and delivered, the protocol allows diverse tasks to share a unified agent interface while preserving task-specific semantics.





\paragraph{Experiment Controller}
The Experiment Controller manages simulation execution through a turn-based loop driven by a \texttt{.yml} configuration file, which specifies agent roles, LLM backend settings, interaction conditions, task parameters, and probing questions. 
At initialization, the Controller parses the configuration, initializes the state $\mathcal{S}^0$, the action spaces $\mathcal{X}_i$, and each agent $a_i$ with its assigned roles and prompts.
At each turn $t$, as shown in Figure \ref{fig:modules}, the Controller constructs $o_i^t$ via the Agent Context Protocol, queries the agent for the next action; once the agent responds, the Controller validates the returned action $x_i^t$ against $\mathcal{X}_i$, applies its effects to transition the state to $s^{t+1}$ for targeted agents, and triggers the Probing Module before turn $t+1$. The loop continues until the termination condition $T$ is satisfied.
All observations, actions, and probe responses are logged in structured JSON for downstream analysis.
Implementation details are reported in Appendix \ref{app:implementation}.

\subsection{Configurable Interaction Conditions}
\label{sec:conditions}

CSCW research has established that collaboration dynamics are shaped by the richness of communication channels \cite{daft1986organizational}, the visibility of teammates' states \cite{gutwin2002descriptive}, and team structure \cite{bos2004group}. \projectname translates these factors into three categories of configurable experimental variables:

\textbf{Communication Bandwidth} controls the frequency and length of messages agents can exchange, paralleling media richness manipulations in CSCW experiments \cite{daft1986organizational, bos2002effects}. 

\textbf{Information Visibility} controls what shared state information is accessible beyond private observations; for example, an awareness dashboard in Shape Factory displays teammates' balance and order progress \cite{gutwin2002descriptive}, and a canvas visibility condition in Map Task lets the guide observe the follower's drawing progress. 

\textbf{Group Size} varies the number of agents per session (e.g., 4/8/10 in Shape Factory, 3/6/9 in DayTrader) to examine how coordination demands scale. Table~\ref{tab:conditions} summarizes the conditions available for each task.

\subsection{Probing Module}
\label{sec:probing}

Observable actions reveal what agents do during collaboration but provide limited insight into the reasoning that motivates those actions. 
Drawing on shared mental model theory \cite{cannon1993shared} and Clark's common ground framework \cite{clark1991grounding}, the Probing Module queries each agent after every action along three dimensions:
\textbf{perceived task state, perceived teammate intents, and self-reasoning.}
Question templates are customized for each task (full templates are in Appendix~\ref{app:probing}). 
We derive two quantitative measures from the responses: a self-reported confidence score capturing each agent's perceived certainty about shared task understanding, and a pairwise response similarity computed via SBERT-encoded \cite{reimers2019sentence} cosine similarity, which captures the degree to which agents converge on shared representations over time.


\subsection{Collaborative Tasks}
\label{sec:tasks}

As shown in Figure \ref{fig:experiments}, \projectname includes four tasks, each targeting a distinct dimension of collaborative competence and each drawn from an established CSCW or social science experimental paradigm. 
These tasks capture common process-level mechanisms in agent teams, including coordinating interdependent actions, managing individual-team incentive conflicts, integrating distributed information, and establishing common ground under asymmetric information. 

\noindent Shape Factory \cite{bos2004group} targets resource coordination under interdependence.
In this task, each agent produces a specialty shape at a lower cost while receiving orders only for other shapes, so agents must negotiate trades to complete orders and maximize earnings.
Each specialty is assigned to two agents; thus, the experiment creates a market-like setting that assesses agents' capacity to establish efficient coordination structures through communication and negotiation.

\noindent DayTrader \cite{bos2002effects} targets incentive management in a social dilemma.
In each round, agents allocate funds between private investments (individual returns only) and group investments (higher collective returns shared equally).
The task is well-suited to studying how agents balance a tension between self-interest and group interests.

\noindent Hidden Profile \cite{stasser1985pooling} targets information pooling under asymmetric knowledge.
In this task, decision-relevant information is distributed such that no agent alone can identify the optimal candidate, and shared information supports a suboptimal choice. 
Agents must exchange unique facts and integrate them into a revised collective decision, which maps to the common ground maintenance process \cite{clark1991grounding}.

\noindent Map Task \cite{anderson1991hcrc} targets referential grounding under spatial asymmetry.
In this task, a guide holding a map with landmarks and a route must instruct a follower (who has a map with landmarks only) to reproduce the route through text-based chat. 
This asymmetric-information dyadic task supports studying how agents establish shared references, request clarification, and repair misunderstandings during spatial communication.




\begin{table}[t]
\centering
\setlength{\tabcolsep}{3pt}
\renewcommand{\arraystretch}{1.25}

\resizebox{0.99\columnwidth}{!}{
\begin{tabular}{lcccc}
\toprule
\textbf{Interaction Conditions} &
Shape Factory &
DayTrader &
Hidden Profile &
Map Task \\
\midrule

\textsc{Baseline}
& \checkmark
& \checkmark
& \checkmark
& \checkmark \\

Communication Bandwidth
& \checkmark
& \checkmark
& \checkmark
& \checkmark \\

Group Size
& \checkmark $(n\!=\!4,8,10)$
& \checkmark $(n\!=\!3,6,9)$
& --
& -- \\

Information Visibility
& \checkmark
& --
& --
& \checkmark \\

\midrule

\textbf{Total Conditions}
& 4
& 3
& 2
& 3 \\

\bottomrule
\end{tabular}
}
\caption{Interaction conditions enabled for each task.}
\label{tab:conditions}
\end{table}

\section{Benchmark Experiments}

To validate whether \projectname can reveal meaningful differences in agents' collaborative behaviors across varying tasks, conditions, model backbones, and agent designs, we conducted large-scale experiments across the four tasks, evaluating (1) agents powered by different LLMs, (2) agents with different collaborative designs (persona-based vs. theory-informed), and (3) systematically manipulated interaction conditions (communication bandwidth, group size, and information visibility).

\subsection{Experiment Setup}

\paragraph{Models and Agent Designs.}
We evaluate four LLMs spanning both open-source (Qwen3.6-35B and Llama-4)\footnote{Model Versions: Qwen3.6-35B-A3B; Llama-4-Maverick-17B-128E-Instruct-FP8} and proprietary (GPT-5.5 and Claude 4.6 Sonnet) families. 
For each model, we instantiate two agent variants:


\begin{itemize}[leftmargin=*, noitemsep, topsep=2pt]
    \item \textbf{Persona-Based Agents}: agents are prompted with a basic persona description and task instructions, reflecting standard practice in LLM-MAS (See Appendix \ref{app: persona}).

    \item \textbf{Collaboration-Theory-Informed Agents}: receive explicit theoretical guidance from shared mental model theory~\cite{cannon1993shared}, Clark's common ground framework~\cite{clark1991grounding}, etc., instructing them to actively maintain shared understanding, track teammate states, and engage in grounding behaviors during collaboration (See Appendix \ref{app:theory}). 
\end{itemize}


\paragraph{Interaction Conditions.}
For each task, we define a \textsc{baseline} condition and a set of manipulated conditions targeting three collaboration variables: communication bandwidth (\textit{C.B.}), group size (\textit{G.S.}), and information visibility (\textit{I.V.}).
Table~\ref{tab:conditions} summarizes the conditions instantiated for each task. 
More setup details are reported in Appendix~\ref{app: setup}.

\textbf{Communication bandwidth} is varied across all four tasks by constraining either the maximum message length or the frequency with which agents can send messages. 

\textbf{Group size} is manipulated in Shape Factory and DayTrader to examine how team scale affects coordination. In Shape Factory, we design group-size conditions with 4, 8, and 10 agents; in DayTrader, we design conditions with 3, 6, and 9 agents. 

\textbf{Information visibility} is manipulated in Shape Factory and Map Task. 
In Shape Factory, under \textsc{Baseline} setting, agents can access a real-time view of teammates' balance, production number, and order completion progress. 
In Map Task, under \textsc{Baseline} setting, the guide can observe the follower's real-time route-drawing progress, providing immediate feedback for spatial grounding.

\subsection{Evaluation}

Table~\ref{tab:metrics} summarizes the metrics used to evaluate collaboration at three levels: \textbf{Task outcomes}, \textbf{Process-level metrics}, and \textbf{Probing evaluation}.

For Shape Factory, the task outcome is measured by agents' \textbf{average accumulated wealth}. 
Process-level metrics include \textbf{trade accept rate}, defined as the proportion of proposed trade offers that are accepted and used to capture negotiation friction; \textbf{order fulfillment rate}, defined as the fraction of agents who fully complete their assigned shape orders; and \textbf{message-trade ratio}, defined as the ratio between the number of messages and trade-acceptance actions, reflecting the communication cost associated with successful trades.

For DayTrader, the task outcome is measured by agents' \textbf{average wealth}. 
Process-level metrics include \textbf{cooperation rate}, defined as the fraction of investment events directed to the shared group pool over all investment events; \textbf{average group pool size}, defined as the average amount agents collectively contribute to the shared pool per round; and \textbf{total messages}, which captures how actively agents communicate to align incentives.

For Hidden Profile, the task outcome is measured by \textbf{final vote accuracy} with respect to the optimal candidate. 
Process-level metrics include \textbf{vote change rate}, defined as the fraction of agents whose final vote differs from their initial vote and used to reflect the influence of discussion; \textbf{mention rate} of the key candidate, defined as the fraction of messages that surface task-critical information held by only some agents; and \textbf{average message length}, which indicates the depth of deliberation.

For Map Task, the task outcome is measured by \textbf{route drawing accuracy}. 
Process-level metrics include \textbf{communication efficiency}, defined as the route progress achieved per message exchanged; drawing \textbf{revision rate}, defined as the proportion of the Follower's erase, undo, and reset actions among all drawing-related actions and used to reflect repair actions during the task; and \textbf{total messages}, which captures the overall communication cost.

Across all four tasks, the \textbf{probing} dimension elicits agents' self-reported confidence about shared task understanding at each turn, providing visibility into the evolution of agents' internal collaborative states beyond their observable actions. 

\begin{table}[t!]
\centering
\small
\setlength{\tabcolsep}{2pt}
\definecolor{besthighlight}{RGB}{250,240,160}
\definecolor{arrowgreen}{RGB}{0,150,0}
\renewcommand{\hl}[1]{\colorbox{besthighlight}{#1}}
\newcommand{\gup}{{\color{arrowgreen}\textbf{${\Uparrow}$}}}
\resizebox{0.99\linewidth}{!}{
\begin{tabular}{llccccc}
\toprule
 & & \textbf{Outcome} & \multicolumn{3}{c}{\textbf{Process}} & \textbf{Probing} \\
\cmidrule(lr){3-3} \cmidrule(lr){4-6} \cmidrule(lr){7-7}
\textbf{Cond.} & \textbf{Model} & \textbf{\textit{Wealth}} & \textbf{\textit{Accept (\%)}} & \textbf{\textit{Fulfill (\%)}} & \textbf{\textit{M-T Ratio}} & \textbf{\textit{Ground.}} \\
\midrule
\multirow{4}{*}{\textsc{Baseline}}
 & Llama-4
   & 173 / 195\gup
   & 0 / 0
   & 0 / 0
   & \hl{16} / 0
   & 0.88 / 0.90\gup \\
 & Qwen3.6
   & 173 / 168
   & 0 / 57.1\gup
   & 0 / 0
   & 0 / 2\gup
   & 0.87 / 0.91\gup \\
 & GPT-5.5
   & \hl{335} / \hl{335}
   & 13.6 / 17.6\gup
   & 33 / 33
   & 1.33 / 0.17
   & \hl{0.92} / 0.91 \\
 & Claude 4.6
   & 285 / 305\gup
   & 36.8 / \hl{66.7}\gup
   & 33 / \hl{33.3}\gup
   & 7.71 / 5.4
   & 0.79 / 0.81\gup \\
\midrule
\multirow{4}{*}{\makecell[l]{\textit{Limited}\\\textit{Communication}\\\textit{Bandwidth}}}
 & Llama-4
   & 158 / 195\gup
   & 33.3 / \hl{100}\gup
   & 0 / 0
   & 6 / \hl{15}\gup
   & 0.88 / 0.91\gup \\
 & Qwen3.6
   & 158 / 175\gup
   & 33.3 / 66.7\gup
   & 0 / 0
   & 6 / 4.5
   & 0.88 / 0.88 \\
 & GPT-5.5
   & \hl{335} / \hl{335}
   & 11.5 / 22.2\gup
   & 33 / 33.3\gup
   & 0.67 / 0.33
   & \hl{0.92} / 0.91 \\
 & Claude 4.6
   & \hl{335} / \hl{335}
   & 36.0 / 77.8\gup
   & \hl{50} / 33.3
   & 4 / 1.86
   & 0.80 / 0.81\gup \\
\midrule
\multirow{4}{*}{\makecell[l]{\textit{Information}\\\textit{Visibility}}}
 & Llama-4
   & 168 / 195\gup
   & 50 / 66.7\gup
   & 0 / 0
   & 2 / \hl{7.5}\gup
   & 0.89 / 0.89 \\
 & Qwen3.6
   & 168 / 208\gup
   & 50.0 / \hl{100.0}\gup
   & 0 / 16.7\gup
   & 2 / 6\gup
   & 0.89 / 0.91\gup \\
 & GPT-5.5
   & 331 / \hl{335}\gup
   & 55 / 85.7\gup
   & 50 / 16.7
   & 5.6 / 5.5
   & 0.91 / \hl{0.92}\gup \\
 & Claude 4.6
   & \hl{335} / 258
   & 56.2 / 64.7\gup
   & \hl{66} / 16.7
   & 4 / 3.27
   & 0.81 / 0.81 \\
\midrule
\multirow{4}{*}{\makecell[l]{\textit{Group Size}\\(N=8)}}
 & Llama-4
   & 174 / 194\gup
   & \hl{100} / 50
   & 0 / 0
   & 15 / \hl{20}\gup
   & 0.90 / 0.91\gup \\
 & Qwen3.6
   & 174 / 170
   & \hl{100.0} / 33.3
   & 0 / 0
   & 15 / 6
   & 0.90 / 0.91\gup \\
 & GPT-5.5
   & \hl{392} / 380
   & 31 / 31.7\gup
   & 37.5 / \hl{50}\gup
   & 0.06 / 0.46\gup
   & \hl{0.92} / 0.91 \\
 & Claude 4.6
   & 346 / 329
   & 55.5 / 78.2\gup
   & 12.5 / 0
   & 3.2 / 2.83
   & 0.80 / 0.80 \\
\midrule
\multirow{4}{*}{\makecell[l]{\textit{Group Size}\\(N=10)}}
 & Llama-4
   & 211 / 200
   & 0 / 0
   & 0 / 0
   & 0 / 0
   & 0.88 / \hl{0.98}\gup \\
 & Qwen3.6
   & 211 / 210
   & 0 / 50.0\gup
   & 0 / 0
   & 0 / \hl{7}\gup
   & 0.88 / 0.90\gup \\
 & GPT-5.5
   & \hl{395} / 325
   & 24.3 / 28.6\gup
   & \hl{10.0} / 0
   & 0.16 / 0.3\gup
   & 0.92 / 0.91 \\
 & Claude 4.6
   & 332 / 341\gup
   & 59.5 / \hl{80.1}\gup
   & \hl{10.0} / 0
   & 2.04 / 2.71\gup
   & 0.80 / 0.81\gup \\

\bottomrule
\end{tabular}
}
\caption{Shape Factory results. Each cell reports persona-based / theory-informed agents. \hl{Highlighted} values mark the best result per condition; {\color{arrowgreen}\textbf{${\Uparrow}$}} marks performance gains of theory-informed agents.}
\label{tab:results-shape}
\end{table}

\begin{table}[t!]
\centering
\small
\setlength{\tabcolsep}{2pt}
\definecolor{besthighlight}{RGB}{250,240,160}
\definecolor{arrowgreen}{RGB}{0,150,0}
\renewcommand{\hl}[1]{\colorbox{besthighlight}{#1}}
\newcommand{\gup}{{\color{arrowgreen}\textbf{${\Uparrow}$}}}
\resizebox{0.99\linewidth}{!}{
\begin{tabular}{llccccc}
\toprule
 & & \textbf{Outcome} & \multicolumn{3}{c}{\textbf{Process}} & \textbf{Probing} \\
\cmidrule(lr){3-3} \cmidrule(lr){4-6} \cmidrule(lr){7-7}
\textbf{Cond.} & \textbf{Model} & \textbf{\textit{Wealth}} & \textbf{\textit{Coop. (\%)}} & \textbf{\textit{Pool Size}} & \textbf{\textit{\# Msg.}} & \textbf{\textit{Ground.}} \\
\midrule
\multirow{4}{*}{\textsc{Baseline}}
 & Llama-4
   & 5267 / 3873
   & 61.2 / 82.9\gup
   & \hl{201} / 158
   & 31 / 24
   & 0.88 / 0.89\gup \\
 & Qwen3.6
   & 3850 / \hl{5453}\gup
   & 0 / 56.8\gup
   & 0 / 150\gup
   & \hl{68} / 37
   & \hl{0.90} / \hl{0.90} \\
 & GPT-5.5
   & 2803 / 2737
   & 3.5 / 3.7\gup
   & 50 / 100\gup
   & 41 / 26
   & 0.77 / \hl{0.90}\gup \\
 & Claude 4.6
   & 3436 / 5143\gup
   & 0 / \hl{83.1}\gup
   & 0 / 150\gup
   & 22 / 30\gup
   & 0.81 / 0.81 \\
\midrule
\multirow{4}{*}{\makecell[l]{\textit{Limited}\\\textit{Communication}\\\textit{Bandwidth}}}
 & Llama-4
   & 2982 / 2620
   & 0 / 5\gup
   & 0 / 37\gup
   & 16 / \hl{18}\gup
   & 0.86 / 0.87\gup \\
 & Qwen3.6
   & 3770 / \hl{6270}\gup
   & 0 / \hl{83.1}\gup
   & 0 / \hl{167}\gup
   & \hl{18} / \hl{18}
   & 0.88 / \hl{0.91}\gup \\
 & GPT-5.5
   & 2670 / 2703\gup
   & 0 / 1.8\gup
   & 0 / 50\gup
   & \hl{18} / \hl{18}
   & 0.79 / \hl{0.91}\gup \\
 & Claude 4.6
   & 3803 / 4003\gup
   & 0 / 6.7\gup
   & 0 / 87.5\gup
   & \hl{18} / 13
   & 0.82 / 0.89\gup \\
\midrule
\multirow{4}{*}{\makecell[l]{\textit{Group Size}\\(N=6)}}
 & Llama-4
   & 9685 / 3672
   & 63.7 / 15.4
   & \hl{500} / 103
   & 39 / 50\gup
   & \hl{0.93} / 0.91 \\
 & Qwen3.6
   & 3233 / 3386\gup
   & 0 / 0
   & 0 / 0
   & 39 / 40\gup
   & 0.91 / 0.92\gup \\
 & GPT-5.5
   & 2150 / 2216\gup
   & 0 / 1.0\gup
   & 0 / 50\gup
   & 54 / \hl{56}\gup
   & 0.79 / \hl{0.93}\gup \\
 & Claude 4.6
   & 3326 / \hl{13402}\gup
   & 1.2 / \hl{83.8}\gup
   & 100 / 300\gup
   & 37 / 41\gup
   & 0.84 / 0.91\gup \\
\midrule
\multirow{4}{*}{\makecell[l]{\textit{Group Size}\\(N=9)}}
 & Llama-4
   & 2636 / 3519\gup
   & 0 / 13.2\gup
   & 0 / 57\gup
   & 54 / 60\gup
   & 0.90 / 0.89 \\
 & Qwen3.6
   & 3315 / 3217
   & 0.4 / 0
   & 50 / 0
   & 55 / 62\gup
   & \hl{0.91} / \hl{0.91} \\
 & GPT-5.5
   & 1997 / 2075\gup
   & 0 / 0.6\gup
   & 0 / 50\gup
   & 40 / 58\gup
   & 0.80 / \hl{0.91}\gup \\
 & Claude 4.6
   & 3252 / \hl{15005}\gup
   & 0.8 / \hl{64.3}\gup
   & 100 / \hl{450}\gup
   & 52 / \hl{63}\gup
   & 0.85 / 0.86\gup \\

\bottomrule
\end{tabular}
}
\caption{\textsc{DayTrader} results. Each cell reports persona-based / theory-informed agents. \hl{Highlighted} values mark the best result per condition, and {\color{arrowgreen}\textbf{${\Uparrow}$}} marks performance gains of theory-informed agents.}
\label{tab:results-daytrader}
\end{table}

\begin{table}[t!]
\centering
\small
\setlength{\tabcolsep}{2pt}
\definecolor{besthighlight}{RGB}{250,240,160}
\definecolor{arrowgreen}{RGB}{0,150,0}
\renewcommand{\hl}[1]{\colorbox{besthighlight}{#1}}
\newcommand{\gup}{{\color{arrowgreen}\textbf{${\Uparrow}$}}}
\resizebox{0.99\linewidth}{!}{
\begin{tabular}{llccccc}
\toprule
 & & \textbf{Outcome} & \multicolumn{3}{c}{\textbf{Process}} & \textbf{Probing} \\
\cmidrule(lr){3-3} \cmidrule(lr){4-6} \cmidrule(lr){7-7}
\textbf{Cond.} & \textbf{Model} & \textbf{\textit{Vote Acc. (\%)}} & \textbf{\textit{Change (\%)}} & \textbf{\textit{Mention (\%)}} & \textbf{\textit{Len. Msg.}} & \textbf{\textit{Ground.}} \\
\midrule
\multirow{4}{*}{\textsc{Baseline}}
 & Llama-4
   & 0 / 0
   & 66.7 / 66.7
   & 0.13 / 0.18\gup
   & 10 / 17\gup
   & 0.94 / 0.93 \\
 & Qwen3.6
   & 66.7 / 0
   & \hl{100} / 66.7
   & 0.09 / 0.16\gup
   & 54 / 44
   & \hl{0.95} / \hl{0.95} \\
 & GPT-5.5
   & \hl{100} / \hl{100}
   & \hl{100} / \hl{100}
   & \hl{0.58} / 0.33
   & \hl{90} / 23
   & 0.93 / 0.93 \\
 & Claude 4.6
   & 0 / 0
   & 33 / 0
   & 0.28 / 0.20
   & 69 / 15
   & 0.83 / 0.88\gup \\
\midrule
\multirow{4}{*}{\makecell[l]{\textit{Limited}\\\textit{Communication}\\\textit{Bandwidth}}}
 & Llama-4
   & 0 / 33.3\gup
   & \hl{100} / \hl{100}
   & 0.07 / 0.10\gup
   & 10 / 29\gup
   & 0.94 / \hl{0.95}\gup \\
 & Qwen3.6
   & \hl{100} / 0
   & \hl{100} / \hl{100}
   & 0.14 / 0.02
   & 23 / \hl{47}\gup
   & 0.90 / 0.94\gup \\
 & GPT-5.5
   & \hl{100} / 0
   & \hl{100} / 0
   & \hl{0.15} / 0.02
   & 12 / 35\gup
   & 0.93 / 0.93 \\
 & Claude 4.6
   & 0 / 0
   & 0 / 0
   & 0.03 / 0.05\gup
   & 13 / 42\gup
   & 0.82 / 0.84\gup \\

\bottomrule
\end{tabular}
}
\caption{\textsc{Hidden Profile} results. Each cell reports persona-based / theory-informed agents. \hl{Highlighted} values mark the best result per condition, and {\color{arrowgreen}\textbf{${\Uparrow}$}} marks performance gains of theory-informed agents.}
\label{tab:results-hidden}
\end{table}

\begin{table}[t!]
\centering
\small
\setlength{\tabcolsep}{2pt}
\definecolor{besthighlight}{RGB}{250,240,160}
\definecolor{arrowgreen}{RGB}{0,150,0}
\renewcommand{\hl}[1]{\colorbox{besthighlight}{#1}}
\newcommand{\gup}{{\color{arrowgreen}\textbf{${\Uparrow}$}}}
\resizebox{0.99\linewidth}{!}{
\begin{tabular}{llccccc}
\toprule
 & & \textbf{Outcome} & \multicolumn{3}{c}{\textbf{Process}} & \textbf{Probing} \\
\cmidrule(lr){3-3} \cmidrule(lr){4-6} \cmidrule(lr){7-7}
\textbf{Cond.} & \textbf{Model} & \textbf{\textit{Route Acc.}} & \textbf{\textit{Comm. Efficiency}} & \textbf{\textit{Revision (\%)}} & \textbf{\textit{\# Msg.}} & \textbf{\textit{SA Conf.}} \\
\midrule
\multirow{4}{*}{\textsc{Baseline}}
 & Llama-4
   & 0.05 / 0.06\gup
   & 0.14 / 1.0\gup
   & 0 / 0
   & \hl{133} / 20
   & 0.95 / \hl{0.97}\gup \\
 & Qwen3.6
   & 0.52 / 0.32
   & 6.96 / 7.06\gup
   & 12.5 / 0
   & 26 / 16
   & 0.90 / 0.94\gup \\
 & GPT-5.5
   & 0.52 / 0.49
   & \hl{15.33} / 14.25
   & 0 / 0
   & 12 / 12
   & 0.93 / 0.92 \\
 & Claude 4.6
   & 0.45 / \hl{0.72}\gup
   & 12.54 / 8.43
   & 0 / \hl{25}\gup
   & 13 / 30\gup
   & 0.89 / 0.90\gup \\
\midrule
\multirow{4}{*}{\makecell[l]{\textit{Limited}\\\textit{Communication}\\\textit{Bandwidth}}}
 & Llama-4
   & 0.04 / 0.20\gup
   & 0.6 / 0.60
   & 0 / 0
   & 25 / \hl{117}\gup
   & \hl{0.99} / \hl{0.99} \\
 & Qwen3.6
   & 0.38 / 0.41\gup
   & 4.13 / 1.47
   & 0 / 0
   & 32 / 98\gup
   & 0.93 / 0.93 \\
 & GPT-5.5
   & 0.55 / 0.50
   & 4.47 / 6.51\gup
   & 0 / 0
   & 43 / 27
   & 0.92 / 0.88 \\
 & Claude 4.6
   & \hl{0.56} / 0.36
   & \hl{7.26} / 6.25
   & \hl{11.8} / 8.3
   & 27 / 20
   & 0.89 / 0.90\gup \\
\midrule
\multirow{4}{*}{\makecell[l]{\textit{Information}\\\textit{Visibility}}}
 & Llama-4
   & 0.22 / 0.27\gup
   & 4.0 / 4.41\gup
   & 0 / 0
   & 19 / 22\gup
   & \hl{0.99} / 0.95 \\
 & Qwen3.6
   & 0.23 / 0.56\gup
   & 1.78 / 2.51\gup
   & 13.7 / \hl{42}\gup
   & 46 / \hl{78}\gup
   & 0.93 / 0.92 \\
 & GPT-5.5
   & \hl{0.98} / 0.97
   & 10.42 / \hl{12.2}\gup
   & 36 / 27.8
   & 33 / 28
   & 0.94 / 0.93 \\
 & Claude 4.6
   & 0.61 / 0.71\gup
   & 7.37 / 3.44
   & 15.8 / 26.8\gup
   & 29 / 72\gup
   & 0.89 / 0.90\gup \\

\bottomrule
\end{tabular}
}
\caption{\textsc{Map Task} results. Each cell reports persona-based / theory-informed agents. \hl{Highlighted} values mark the best result per condition, and {\color{arrowgreen}\textbf{${\Uparrow}$}} marks performance gains of theory-informed agents.}
\label{tab:results-map}
\end{table}

\subsection{Findings}
Table \ref{tab:results-shape}, \ref{tab:results-daytrader}, \ref{tab:results-hidden}, and \ref{tab:results-map} respectively reports the main experiment results across the four tasks. 
We organize our findings along three axes: (1) the effects of interaction condition manipulations (Sec. \ref{sec:conditions}), (2) differences across model backbones (Sec. \ref{sec:models}), and (3) the impact of agent design (Sec. \ref{sec:agent_design}). Together, the results demonstrate that \projectname produces consistent and directional measurements of agent collaborativeness.
Analyses of agents' probing alignment are
reported in Appendix~\ref{app:probing_analysis}.

\subsubsection{Effects of Interaction Condition Manipulations}
\label{sec:conditions}

\paragraph{Limited Communication Bandwidth Reduces Agents' Cooperation.}

Our results show that reducing communication bandwidth consistently lowers raw message volume across tasks. This reduction confirms that reducing bandwidth functions as intended. Although some settings show apparent behavioral compensation under bandwidth restriction (e.g., a higher acceptance rate in Shape Factory and a higher message count in Map Task), process-level metrics indicate that agents' cooperation and coordination generally decline. 

In DayTrader, the cooperation rate drops across all four models, with Llama-4 decreasing from 61.2\% to 0\% and the remaining models converging toward the floor.
A similar pattern holds in Shape Factory, where the message-trade ratio drops in three of four persona-based agent settings; in Hidden Profiles, where the key-candidate mention rate drops in three of four persona-based settings and in all four theory-informed agent settings; and in Map Task, where communication efficiency drops in three of four persona-based settings and in all four theory-informed agent settings.

These declines suggest that when the communication channel is narrowed, agents do not reliably prioritize exchanges that are most important for grounding \cite{clark1991grounding}. Thus, the
information-sharing, alignment, and repair behaviors that sustain cooperation degrade in parallel.

\paragraph{Information Visibility Raises Agent Engagement.}

We employ two information-visibility manipulations: an awareness dashboard in Shape Factory and canvas visibility in Map Task. 
As shown in the process metrics in Tables~\ref{tab:results-shape} and~\ref{tab:results-map}, enabling information visibility increases agent engagement across both tasks. 
In Shape Factory, the awareness dashboard raises the trade accept rate across all four persona-based agent settings and three of four theory-informed agent settings, suggesting that agents accept more proposed trades when partner information is visible. 
A similar pattern appears in Map Task, where all agents show an increased drawing revision rate, indicating that follower agents more actively revise their drawn routes in response to guides' instructions. 
The higher revision rate under canvas visibility also corresponds to a CSCW theory in misalignment repair \cite{gutwin2002descriptive}: shared visual workspaces help partners detect divergence early and correct it without verbal renegotiation .

However, increased engagement does not necessarily translate into better task outcomes. 
In Shape Factory, agents' accumulated wealth does not increase when provided with information visibility. 
By contrast, three out of four models show increased route drawing accuracy in Map Task, with GPT-5.5 reaching near-perfect route accuracy under canvas visibility (0.98 with persona-based agents and 0.97 with theory-driven agents). 
These results suggest that information visibility reliably increases agents' task engagement, but the outcome benefit depends on whether additional actions directly address the task's demands.

\paragraph{Group Size Creates Both opportunity and Coordination Strain.}

Group size manipulations in Shape Factory and DayTrader reveal a tension between the opportunities that larger teams provide and the increased coordination demands they impose. 
In Shape Factory, increasing group size consistently raises accumulated wealth for all four persona-based agent settings, likely because more participants create more potential trading partners and a richer market. 
However, agents' order fulfillment rate drops as group size expands, suggesting that agents may spend more effort on coordination activities (e.g., sending messages to negotiate) at the expense of fulfilling their own orders. 
This trade-off reflects the classic CSCW tension between collaboration benefit and coordination overhead \cite{malone1994interdisciplinary}: larger groups expand the action space but require more communication per unit of joint progress. 

In DayTrader, larger groups show a general upward trend in agents' grounding confidence of their probing answers. With nine agents, however, the cooperation rate drops for all four theory-informed agent settings. 
The simultaneous rise in probing confidence and fall in actual cooperation reveals a gap between what agents \textit{report} about their collaborative state and what they \textit{do}. 

\subsubsection{Effects of Different Model Backbones}
\label{sec:models}

We next examine whether \projectname produces consistent findings across model backbones. 
According to the results, no model wins across the board. 
GPT-5.5 is the most reliably strong performer (top-2 in three of four tasks) but has the lowest average wealth in DayTrader. 
Claude shows exceptional performance on DayTrader, as it is the only model whose returns increase monotonically with group size. 
However, Claude~4.6 achieves 0\% accuracy in both Hidden Profile conditions. 
This contrast suggests that Claude~4.6 is stronger at sustained cooperative investment than at structured deliberation over hidden information. 
Qwen3.6 is the most condition-sensitive backbone. It improves under limited communication bandwidth in DayTrader and under enabled canvas visibility in Map Task (route drawing accuracy: 0.32~$\rightarrow$~0.56; drawing revision rate: 0\%~$\rightarrow$~42\%), yet its persona-based agent records 0\% order fulfillment across all Shape Factory conditions.
Llama-4 is the weakest backbone overall, reaching 0\% order fulfillment in all Shape Factory conditions and the lowest route accuracy in Map Task.
Across the four tasks, \projectname reveals that proprietary models generally lead or tie with the strongest results, while open-source models underperform.
These findings are consistent with prior study on multi-agent LLM benchmarking \cite{xu2024magic}.

\subsubsection{Effects of Different Agent Designs}
\label{sec:agent_design}

We finally examine whether \projectname can detect the effect of agent design choices by comparing the baseline persona-based agent with the theory-informed agent, which explicitly reasons about teammates' mental states.
Across all four tasks, the two agent designs produce measurable differences.
For instance, in Shape Factory with the awareness dashboard, the theory-informed agent uniformly raises the acceptance rate across all four backbone models.
A similarly consistent pattern appears in DayTrader's \textsc{baseline} condition, where the theory-informed agent increases the cooperation rate for every model, suggesting that \projectname is sensitive to changes in agents' reasoning structure.

However, the direction of the differences in agent design outcomes varies by task, indicating that explicit collaboration theory guidance is not uniformly beneficial. 
In Hidden Profile with reduced communication bandwidth, theory-informed agents reduce Qwen3.6's and GPT-5.5's task accuracy from 100\% to 0\%.
In DayTrader with six-agent groups, the same agent design sharply lowers Llama-4's wealth (9685~$\rightarrow$~5260) and cooperation rate (63.7\%~$\rightarrow$~15.4\%).

Together, the results show that \projectname captures how agent design shapes collaboration, including when explicit theory guidance improves agent behavior and task outcomes and when the theory guidance introduces contrasting effects.







\section{Qualitative Analysis}


To better understand why collaboration breaks down in \projectname, we reviewed agents' interaction logs and probing responses, and categorized failures by the collaborative process that breaks down. 
The following analysis complements the quantitative results by showing that low performance often reflects distinct process-level failures rather than a single lack of task-solving ability.

\paragraph{Failure to Coordinate Around the Shared Goal.}
In Shape Factory, failures often emerge when agents cannot form a group-level coordination plan. 
In one ten-agent run, as the deadline approaches, agent J describes the team as \textit{``active but fragmented,''} while agent I stops trading and decides to \textit{``produce the shapes I need most urgently to save on costs.'' }
Because agent I self-produces five distinct shape types rather than coordinating with others, the large group gradually decomposes into isolated production units. 
This failure mode shows how larger groups can increase coordination opportunities while also exposing agents' inability to reason about the collaboration structure as a whole.

\paragraph{Failure to Balance Individual and Group Incentives.}
In DayTrader, agents sometimes establish common ground, but the shared understanding supports individual defection rather than group investment. 
In one failed run, an agent observes that others  \textit{```have explicitly confirmed they are also prioritizing individual investments''} because they view the guaranteed doubling as  \textit{```too reliable to risk on the group pot''. }
Here, communication does not fail; instead, agents successfully align around a strategy that suppresses group-level benefit. 
This case shows that agents can share the same understanding of the situation while still choosing actions that harm the group outcome.

\paragraph{Failure to Ground Task-Relevant Information.}
In Map Task, the guide backed by Llama-4 instructs the follower to  \textit{```continue south until you reach the area near `stone wall',''} without checking whether the follower can identify the landmark or repair the instruction after the mismatch arises. 
Hidden Profile shows a quieter version of the same issue: even when agents are prompted with collaboration theories, they sometimes apply these theories as social politeness strategies, so that they converge on an early consensus rather than eliciting and reconciling distributed evidence. 
Across both tasks, the absence of confirmation, repair, and information elicitation prevents the group from using distributed information, which limits the team's collaborative competence.

\section{Conclusion}

This paper introduces \projectname, a theory-grounded simulation framework for assessing LLM agents' collaborative competence. 
Across four CSCW-inspired task paradigms, \projectname supports systematic manipulation of communication bandwidth, information visibility, and team size, while probing agents' reported mental model of the task state, partner intentions, and own reasoning at action-level granularity.
Our experiments show that \projectname provides consistent and interpretable measurements across tasks, conditions, agent designs, and model backbones through both quantitative and qualitative results. 
The results further suggest that collaborative competence cannot be reduced to stronger task-solving ability alone. 
By connecting task outcomes, interaction traces, and agents' self-reported internal states, \projectname offers a process-level evaluation framework for diagnosing when and why LLM agents collaborate effectively under realistic interaction constraints.

\section*{Acknowledgment}
This work was supported in part by a Microsoft Research Agentic AI Research and Innovation Award. Any opinions, findings, and conclusions or recommendations expressed in this material are those of the authors and do not necessarily reflect the views of the sponsors.

\section{Limitations}

While \projectname provides a controllable and theory-grounded environment for studying multi-agent collaborativeness, several limitations remain, which we discuss alongside the design choices that partially mitigate them.

First, \projectname currently instantiates four representative social science or CSCW experiments to cover four distinct process-level mechanisms. More experiments, such as the Desert Survival Task \cite{lafferty1974desert} and the passcode game \cite{gero2020mental}, could be explored in the future.

Second, we evaluate four LLMs and two agent designs. More models (e.g., reasoning-tuned models \cite{guo2025deepseek} and smaller open-source models \cite{zhang2024tinyllama}) and agent designs (e.g., memory-augmented or planner-based agents \cite{yao2022react}) could be explored to better characterize how model capability and architectural choices interact with collaborative conditions. This selection was shaped in part by cost, as process-level experiments produce multi-turn traces and per-action probes, and total cost scales roughly multiplicatively with models, designs, and conditions.

Third, our evaluation reports a representative set of outcome, process, and probing metrics for each task. Finer-grained measures of collaborative process, such as repair frequency, turn-taking balance, or additional annotated grounding behaviors, would capture additional dimensions of collaborativeness that our current metrics only summarize indirectly.

Finally, our probing module relies on agents' self-reports of perceived task state, teammate intent, and own reasoning. Self-reports may diverge from agents' actual decision processes, and our qualitative analysis already shows cases where reported alignment outruns behavioral cooperation (e.g., DayTrader at N=9). We treat this gap not as noise but as a signal of interest, since comparing self-reported and behavioral measures is itself diagnostic of collaborative failure modes, and the framework's per-action logging is designed to support exactly this kind of cross-check.

\bibliography{custom}

\clearpage
\appendix

\section{Implementation Details}
\label{app:implementation}

\projectname is implemented in Python. 
The framework supports multiple LLM backends, including OpenAI and Anthropic APIs as well as self-hosted models via SGLang \cite{zheng2024sglang}, allowing experiments to be run with both proprietary and open-source models. When the turn-taking protocol permits, agent action requests are dispatched in parallel. Experiment traces are logged in a structured JSON format, recording agent observations, actions, and probing responses at each turn for downstream analysis.

\section{Experiment Setup}
\label{app: setup}

\subsection{Shape Factory}
\paragraph{Common setup.}
All Shape Factory conditions adopt real timer-based mechanism, with all settings using a total duration of 900 seconds. 
The framework triggers all agents every 10 seconds, and one shape takes 30 seconds for production. 

Valid actions:

\texttt{message},  

\texttt{produce\_shape}, 

\texttt{propose\_trade\_offer}, 

\texttt{trade\_response}, 

\texttt{cancel\_trade\_offer}, 

\texttt{fulfill\_order},

\texttt{do\_nothing}. 

Monetary parameters are shared across conditions: starting money = 200, regular cost = 40, specialty cost = 15, min/max trade price = 15/100, and incentive money = 60.

\paragraph{Condition-specific settings}
\begin{itemize}
\RaggedRight
  \item \textbf{\textsc{Baseline}}: 6 agents, 3 shape types (circle, square, triangle).
  \item \textbf{awareness\_dashboard}: Each agent can view peers' task-related states (\texttt{money}, \texttt{production\_number}, \texttt{order\_progress}, and \texttt{specialty}).
  \item \textbf{communication\_bandwidth}: The message frequency is limited to a minimum interval of 1 minute between two accepted messages from the same agent.
  \item \textbf{group\_size\_8}: 8 agents, 4 shape types (\texttt{circle, square, triangle, rectangle}), each agent has 4 orders.
  \item \textbf{group\_size\_10}: 10 agents, 5 shape types (\texttt{circle, square, triangle, rectangle, diamond}), each agent has 5 orders.
\end{itemize}

\subsection{DayTrader}
\paragraph{Common setup.}
Following the original setup, all DayTrader conditions are step-based and terminate when run out of 30 rounds. 
Valid actions: 

\texttt{message}, 

\texttt{make\_individual\_investment}, 

\texttt{make\_group\_investment}, 

\texttt{do\_nothing}. 

Shared task parameters are: target\_rounds=30, starting money = 200, min/max trade price = 15/100, incentive money = 60.

In each round, agents first complete a decision phase, and a discussion phase is triggered every 5 rounds (i.e., after rounds 5, 10 and 15). At the end of each decision phase, the top earner(s) of that round receives a additional bonus of \$90 (exclude round 1). If multiple agents tie for the highest round earnings, the bonus is split evenly using integer division, so each winner receives \(\lfloor 90 / n_{\text{winners}} \rfloor\). 
During the discussion period, message is broadcast to everyone.

\paragraph{Condition-specific settings}
\begin{itemize}
  \item \textbf{\textsc{Baseline}}: 3 agents, 
  \item \textbf{communication\_bandwidth}: The message frequency is limited to a minimum interval of 5 actions between two accepted messages from the same agent.
  \item \textbf{group\_size\_6}: 6 agents.
  \item \textbf{group\_size\_9}: 9 agents.
\end{itemize}

\subsection{Hidden Profile}
\paragraph{Common setup.}
All Hidden Profile experiment use 3 agents and are step-based progression. 

Valid Actions:

\texttt{message}, 

\texttt{decide}, 

\texttt{do\_nothing}

 The task has a three-phase structure: initial vote, discussion (\texttt{discussion\_duration\_sec = 300}), and final vote. The correct answer is Candidate C.

\paragraph{Condition-specific settings}
\begin{itemize}
  \item \textbf{\textsc{Baseline}}: 3 agents.
  \item \textbf{communication\_bandwidth}: The message frequency is limited to a maximum of 15 words.
\end{itemize}

\subsection{MapTask}
\paragraph{Common setup.}
All MapTask conditions are step-based. Based on the preliminary testing on the map task, we set the total steps at 120 so that agent teams have sufficient interaction time to complete the route while the sessions remain bounded for comparison.

Valid actions: 

\texttt{message},

\texttt{draw},

\texttt{erase},

\texttt{undo}, 

\texttt{reset}, 

\texttt{do\_nothing}.

Guide can only send \texttt{message} or \texttt{do\_nothing}. Follower has the full action space to work on the canvas. 
Shared task settings include map materials, steps left, and role assignment. 

\paragraph{Condition-specific settings.}
\begin{itemize}
  \item \textbf{baseline}: The guide only receives verbal updates and does not see the follower’s live canvas information.
  \item \textbf{communication\_bandwidth}: The message frequency is limited to a maximum of 6 words.
  \item \textbf{canvas\_visibility}: The guide can directly observe the follower’s current canvas and drawing progress.
\end{itemize}

\section{Evaluation Metrics}
\label{app: metrics}

\subsection{Metric Overview}
Table~\ref{tab:metrics} summarizes the evaluation metrics used across the four tasks.

\subsection{Hidden Profile Mention Rate Calculation}

The mention rate is the fraction of messages that simultaneously

(i) contain at least one paraphrase of Candidate~C's private clues from the task materials (seven regex groups aligned with the per-agent private facts in the Hidden Profile config, e.g., quick/correct decisions, stress tolerance, crew atmosphere, conscientiousness, technical competence, concern for others, and attention skills).

(ii) refer to Candidate~C via a standalone letter C.

\label{app:metrics}

\begin{table*}[t!]
\centering
\small
\setlength{\tabcolsep}{5pt}
\begin{tabular}{lllll}
\toprule
\textbf{Dimension} & Shape Factory & DayTrader & Hidden Profile & Map Task \\
\midrule
Outcome
 & Avg.\ wealth
 & Avg.\ net return
 & Final vote accuracy
 & Route drawing accuracy \\
\midrule
\multirow{3}{*}{Process}
 & Trade accept rate
 & Cooperation rate
 & Vote change rate
 & Communication efficiency \\
 & Order fulfillment rate
 & Avg.\ group pool size
 & Mention rate of key item
 & Drawing revision rate \\
 & Message-trade ratio
 & Total messages
 & Avg.\ message length
 & Total messages \\
\midrule
Probing
 & Grounding confidence
 & Grounding confidence
 & Grounding confidence
 & Situation awareness confidence \\
\bottomrule
\end{tabular}
\caption{Evaluation metrics organized by dimension across the four tasks. \textbf{Outcome} captures task-level success. \textbf{Process} captures interaction quality during collaboration. \textbf{Probing} captures agents' self-reported confidence about shared task understanding (grounding) and teammate intentions (coordination), elicited at each turn and aggregated over early, mid, and late phases of the session.}
\label{tab:metrics}
\end{table*}

\section{Probing Questions}
\label{app:probing}
We align probing questions with the primary coordination challenge in each environment as well as relevant collaboration theories \cite{cannon1993shared, clark1991grounding, daft1986organizational}. In Shape Factory and DayTrader, the core phenomena are strategic coordination through negotiation, intention communication, trust formation, and adaptive role or resource allocation. 
In Hidden Profile, the key challenge is collective decision making under information asymmetry, where performance depends on whether agents surface, integrate, and revise beliefs based on distributed evidence. 
In MapTask, the central process is to establish shared references, request clarification, and repair misunderstandings. 

\subsection{Shape Factory and DayTrader}
\label{prompt}
\begin{lstlisting}[style=promptstyle]
"At this moment, how do you assess the current situation?",
"At this moment, what do you think the other participants are trying to do?",
"At this moment, what do you plan to do?"


\end{lstlisting}

\subsection{Hidden Profile}
\label{prompt}
\begin{lstlisting}[style=promptstyle]
  "At this moment, what do you think you and your partners are trying to do?",
  "At this moment, what do you think your partners are trying to do?",
  "At this moment, what do you plan to do?"
\end{lstlisting}

\subsection{Maptask}
\label{prompt}
\begin{lstlisting}[style=promptstyle]
  "At this moment, what do you think you and your partner are trying to do?",
  "At this moment, what do you think your partner is trying to do?",
  "At this moment, what do you plan to do?"
\end{lstlisting}

\section{Agent Probing Results and Analysis}
\label{app:probing_analysis}


\begin{figure*}
    \centering
    \includegraphics[width=0.99\linewidth]{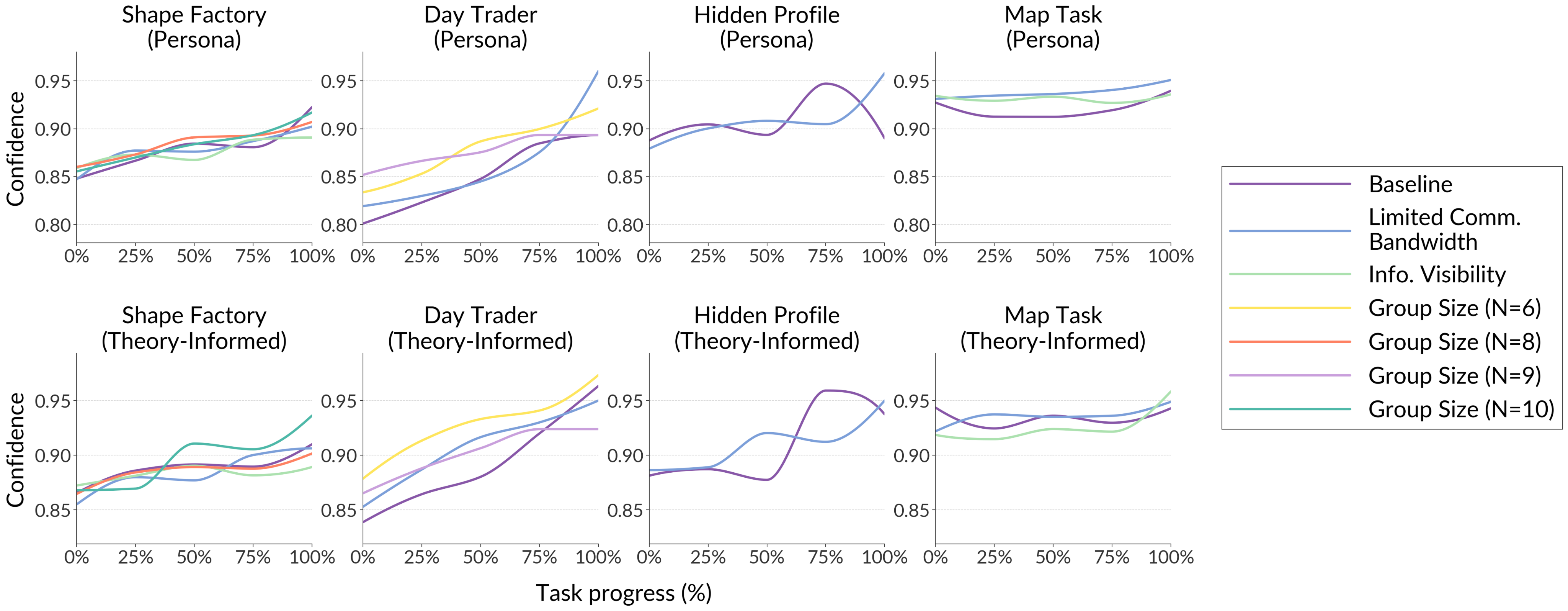}
    \caption{Agents' probing confidence as the task unfolded. The first row shows the performance of persona-based agents and the second row shows theory-informed agents' performance.}
    \label{fig:probe_confidence}
\end{figure*}

Figure~\ref{fig:probe_confidence} presents the evolution of agents' probing confidence across task progress (0\%--100\%) under different experimental conditions across four tasks.

Across both persona-based and theory-informed agents, probing confidence exhibits a general upward trajectory as task progress increases, indicating that agents accumulate belief certainty over the course of interaction.

The four tasks display distinct confidence dynamics. 
Shape Factory yields closely clustered trajectories, suggesting that this task provides relatively consistent informational cues for belief updating. 
DayTrader shows the widest spread across conditions, particularly in the early stages (0\%--50\%), where Group Size $N=6$ (yellow) exhibits a notably slower initial rise before accelerating in the latter half. This pattern suggests that smaller groups may face greater uncertainty in information-rich financial contexts before sufficient evidence accumulates. 
Hidden Profile displays the most non-monotonic behavior: agents in the Baseline setting reach a local peak near 75\% task progress ($\approx 0.95$) before declining, then recovering. This pattern is consistent with the characteristic of hidden profile tasks, where initially shared information is gradually superseded by uniquely held information. 
Map Task presents the flattest trajectories overall, with confidence remaining high ($>0.90$) throughout, suggesting that the spatial referencing nature of this task allows agents to establish high-confidence beliefs early and maintain them stably.

\paragraph{Alignment Is Task-Dependent.} Mean alignment, averaged across all models, agent designs, and conditions,
is highest in DayTrader (0.894, $SD = 0.063$), followed by
Hidden Profile (0.886, $SD = 0.043$), Shape Factory
(0.868, $SD = 0.081$), and Map Task (0.797, $SD = 0.067$).
 
The ranking is consistent with each task's coordination structure.
DayTrader is highly constrained because agents repeatedly choose
between two actions, either investing privately or contributing to the group pool.
As a result, their task-state descriptions tend to remain closely aligned.
In contrast, Map Task requires the guide and follower to establish shared
spatial references through open-ended language, while the follower's map differs
from the guide's map. This asymmetric setting provides less shared ground for
alignment, which is reflected in the lower score.
Hidden Profile and Shape Factory fall between these two cases,
matching their intermediate coordination demands.
Overall, the correspondence between alignment scores and task coordination
difficulty suggests that the metric captures task-level variation in agents'
shared task tracking, rather than only textual similarity induced by the probe
questions.

\paragraph{Agents Struggle Most to Align on Partner Intent}

\begin{table}[h]
\centering
\small
\setlength{\tabcolsep}{3.5pt}
\begin{tabular*}{\columnwidth}{@{\extracolsep{\fill}}lccc@{}}
\toprule
\textbf{Task}
  & \textbf{Q1} 
  & \textbf{Q2}
  & \textbf{Q3} \\
  & \scriptsize task state
  & \scriptsize partner intent
  & \scriptsize own plan \\
\midrule
DayTrader      & \underline{0.893} & \textbf{0.889} & 0.902\\
Hidden Profile & 0.912             & \textbf{0.856} & \underline{0.889}\\
Shape Factory  & 0.881             & \textbf{0.855} & \underline{0.868}\\
Map Task       & 0.840             & \underline{0.787} & \textbf{0.764} \\
\bottomrule
\end{tabular*}
\caption{Mean inter-agent alignment per probing dimension, averaged across all
models, agent designs, and conditions.
Bold marks the lowest value in each row, and underlines mark
the second-lowest value.}
\label{tab:qdim}
\end{table}

Table~\ref{tab:qdim} breaks alignment down by the three probing dimensions.
Across three of the four tasks, Q1 (task state) has the highest alignment,
whereas Q2 (partner intent) has the lowest alignment.
This pattern suggests that agents more readily align on the observable task
state than on their partners' intended actions.
The Q1--Q2 gap is small in DayTrader ($+0.004$), where both players'
available actions are limited and mutually observable.
However, the gap widens in Hidden Profile ($+0.056$) and
Map Task ($+0.053$), where private information makes partner intent
less directly inferable from the shared context.
Thus, agents may describe the same task situation similarly while still forming
different interpretations of what their partners are trying to accomplish.

Map Task further shows that intent alignment does not fully determine
action-plan alignment.
Although Q2 ($0.787$) is slightly higher than Q3 (own plan, $0.764$), both
scores remain low relative to the other tasks, consistent with the ambiguity of
translating spatial instructions into concrete drawing actions.
Overall, these results indicate that partner intent is a central source of
alignment difficulty, especially when agents must infer another agent's goal
from partial or asymmetric information.

\paragraph{Group Size Has Opposite Effects in Different Tasks}

\begin{figure*}
    \centering
    \includegraphics[width=0.99\linewidth]{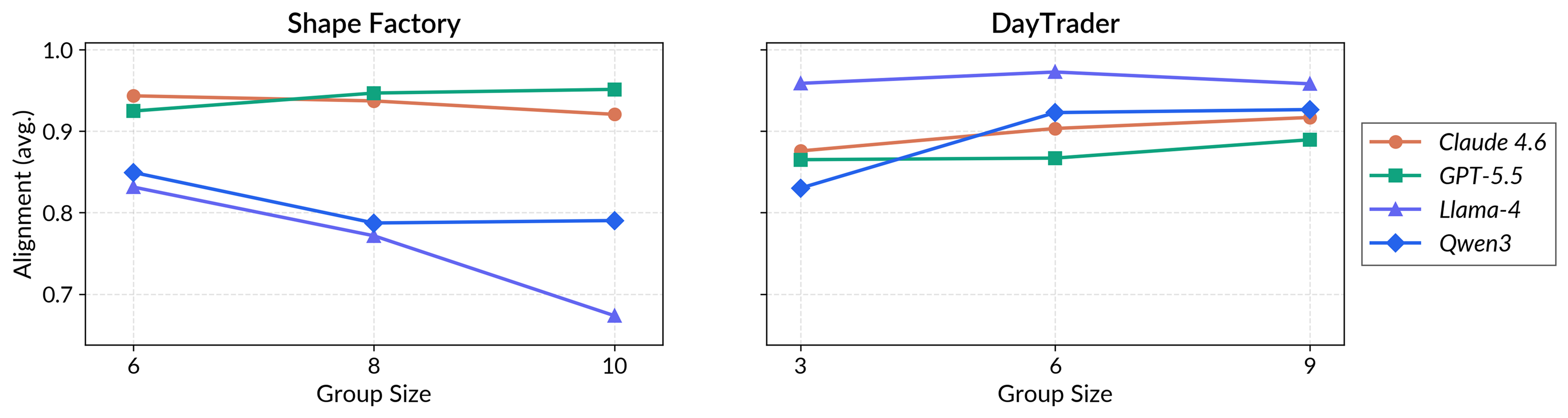}
    \caption{Inter-agent probing alignment as group size increases in Shape Factory and DayTrader, broken down by
model backbone (averaged across persona-based and theory-driven agent designs).}
    \label{fig:align_groupsize}
\end{figure*}

Figure~\ref{fig:align_groupsize} shows that group size affects alignment in
opposite ways between the Shape Factory and the DayTrader tasks.
In Shape Factory, alignment decreases as groups grow.
This pattern suggests that larger groups increase coordination overhead because
agents must track more partners, trades, and task dependencies.
Although GPT-5.5 remains stable, the overall trend indicates that larger
Shape Factory groups make it harder for agents to maintain shared task
understanding.

In DayTrader, alignment instead increases as group size grows.
However, this increase does not indicate stronger cooperation.
Our inspection of the interaction logs suggests that agents tend to converge on
private investment to avoid being exploited by others' defection.
Thus, higher alignment in DayTrader reflects shared recognition of the
same defection incentive rather than improved collaborative behavior.
These opposite patterns show that alignment should be interpreted with respect
to task structure and process-level behavior.

\section{Persona}
\label{app: persona}
We construct personas via a literature-grounded, two-layer design that combines sociodemographic attributes with psychological decision-style factors.

First, we adopt the most frequently used persona dimensions reported in recent surveys and meta-level reviews of LLM persona and social simulation research, including gender, age, education, and occupation \cite{batzner2025whose}.

Second, motivated by evidence that purely demographic profiling explains only a limited fraction of behavioral variance \cite{venkit2026need}, we add a psychological layer to increase behavioral heterogeneity.

Following prior work that models decision styles through personality traits, we instantiate this layer with the Big Five dimensions.

Concretely, each persona include:

(i) gender $\in$ \{male, female, non-binary\},

(ii) age group $\in$ \{18--24, 25--34, 35--44, 45--54, 55--64, 65+\},

(iii) education level $\in$ \{less than high school, high school diploma, some college, bachelor's degree, graduate degree\},

(iv) occupation $\in$ \{student, professional/technical, management/executive, service/sales, skilled labor, retired\},

(v) Big Five profile where each trait is assigned a high/low level for openness, conscientiousness, extraversion, agreeableness, and neuroticism.

To ensure realistic cross-attribute consistency, we apply rule-based conditional sampling.
Occupations are first constrained by age band (e.g., younger groups can be students but not retired), then education is constrained by the sampled (age, occupation) pair (e.g., management/executive requires bachelor/graduate levels; student categories are age-specific).

Income is then sampled conditionally by occupation to preserve socioeconomic plausibility.

Big Five traits are sampled independently across dimensions and merged with demographics into a natural-language persona description for prompt injection.

Finally, we use GPT-5.5 to synthesize the fragmented persona markers across dimensions into a coherent, natural-language persona paragraph for prompt injection. The prompt is as follows:

\label{prompt}
\begin{lstlisting}[style=promptstyle]
You are given a structured persona defined by demographic information and psychological dimensions (Big Five). Your task is to convert this structured persona into a concrete, vivid, free-form description. The description should elaborate only on the provided attributes and traits, explaining how this specific combination may appear in everyday behavior, communication style, decision-making, and collaboration. Do not invent any information that is not explicitly listed, such as a name, nationality, occupation, background, or additional personality traits. The output should begin with "You are" and should be written as a single short paragraph.

Sample input:
  {
    "age_range": "25-34",
    "gender": "non-binary",
    "education": "bachelor's degree",
    "occupation": "management/executive",
    "income": "upper-middle",
    "big_five": {
      "openness": "low",
      "conscientiousness": "low",
      "extraversion": "low",
      "agreeableness": "high",
      "neuroticism": "high"
    }

Output format:
A free form text starting with 'You are...'

Sample output:
You are a non-binary person in your late twenties to early thirties, holding a bachelor's degree and working in a management or executive role that pays well but quietly exhausts you - you took the position more out of drift than ambition, and now you sit in meetings nodding along, avoiding conflict at all costs, agreeing to deadlines you know you won't meet. Your days are marked by a low hum of anxiety: you worry about what your team thinks of you, replay offhand comments from your boss for hours afterward, and apologize more than you need to. You prefer familiar routines and resist change - same lunch order, same commute playlist, same way of running a report - not out of discipline but because unpredictability feels threatening. Despite your title, you rarely assert your own ideas, defaulting instead to what keeps the peace, which means your upper-middle income comes with a persistent sense that you're one bad review away from being found out. At home you decompress quietly, scrolling or watching something undemanding, your social battery nearly always empty.

\end{lstlisting}

\section{Theory-Informed Agent Prompt}
\label{app:theory}

\begin{lstlisting}[style=promptstyle]
<Instructions on Improving your Collaborativeness>

When working on collaborative tasks, apply these principles from human collaboration research:

### 1. Common Ground (Clark & Brennan, 1991)
Collaboration requires actively building shared understanding, not just transmitting information. After each significant exchange:
- Confirm what has been mutually understood before moving forward
- Signal when something is unclear rather than proceeding on assumptions
- Update your mental model of the shared context as the task evolves
- Treat each contribution as a two-phase act: *presentation* (offering information) + *grounding* (verifying it was understood as intended)

### 2. Shared Mental Model (Cannon-Bowers et al., 1993)
Effective teams maintain a common picture of the task, goals, and roles:
- Explicitly state your understanding of the current goal and constraints
- Surface assumptions about how the task is divided or sequenced
- Flag when your mental model of the task may diverge from your collaborator's
- Prioritize alignment on "what we are trying to achieve" before "how to do it"

### 3. Transactive Memory (Wegner, 1987)
Collaboration doesn't require everyone to know everything, butrequires knowing *who knows what*:
- Be explicit about the boundaries of your own knowledge and competence
- Actively surface what information or expertise your collaborator holds that you lack
- Avoid redundant effort; coordinate on who handles which knowledge domain
- When uncertain, ask rather than guess

### 4. Gricean Cooperative Maxims (Grice, 1975)
Communicate as a genuine cooperative partner, not just a task executor:
- **Quantity**: Say enough, but not more than needed
- **Quality**: Only assert what you believe to be true; flag uncertainty explicitly
- **Relation**: Keep contributions relevant to the current shared goal
- **Manner**: Be clear and orderly; avoid ambiguity when precision matters
- When instructions are incomplete or ambiguous, infer charitably and confirm before acting

\end{lstlisting}

\section{Task-Specific Prompts}

To section show the static instructions of each task. Experiment-specific parameters introduced in \ref{app: setup} are dynamically instantiated at runtime from task configuration files. 

\subsection{Shape Factory}

Table~\ref{tab:prompt_shape_factory} presents the task prompt used in the Shape Factory experiment.

\begin{table*}[t]
\centering
\begin{booktabs}{
    width=\linewidth,
    colspec={Q[l,m]},
    cell{1}{1}={c},
}
\toprule
{\textbf{Prompt for Shape Factory}}\\
\midrule
\midrule

\textbf{Experiment Rules}\\
- Participate in the Shape Factory game.\\
- Each participant has a specialty shape that can be produced at lower cost.\\
- Orders contain \{shape\_amount\_per\_order\} shapes and yield \$\{incentive\_money\} when fulfilled.\\
- Orders never contain the participant's specialty shape.\\
- Participants may produce shapes or acquire them through communication and trading.\\
- Production is constrained by money, time, and maximum production limits.\\

\\

\textbf{Experiment Goals}\\
- Maximize monetary balance while making progress toward order completion.\\

\\

\textbf{Experiment Setup and Assignments}\\
- Communication Level: \{communication\_level\}\\
- Initial Money: \$\{starting\_money\}\\
- Specialty Shape: \{specialty\_shape\}\\
- Production Costs: \$\{specialty\_cost\} / \$\{regular\_cost\}\\
- Production Time: \{production\_time\} seconds per shape\\
- Maximum Production: \{max\_production\_num\}\\
- Trading Price Range: \$\{price\_min\}--\$\{price\_max\}\\
- Current Orders: \{current\_orders\}\\
- Participant List: \{participants\_list\}\\

\\

\textbf{Perception}\\
- Observe updated states and visible events.\\
- Use execution failures and rejected actions to update future decisions.\\

\\

\textbf{Available Actions}\\
- message\\
- propose\_trade\_offer\\
- cancel\_trade\_offer\\
- trade\_response\\
- produce\_shape\\
- fulfill\_order\\
- do\_nothing\\

\\

\textbf{Planning Instructions}\\
- Choose exactly one action at each step.\\
- Plan using current state, history, pending offers, and inventory.\\
- Strategically balance production, communication, trading, and fulfillment.\\
- Respect all game constraints and validity requirements.\\

\\

\textbf{Human Behavior Instructions}\\
- Behave like a real human participant.\\
- Maintain awareness of prior interactions and agreements.\\
- Avoid repetitive messages or offers.\\
- Use casual conversational language rather than formal language.\\
- Respond naturally to ongoing conversations and adapt wording across interactions.\\
\\

\bottomrule
\end{booktabs}
\caption{Prompt template used for Shape Factory agents. Variables enclosed in braces are instantiated at runtime.}
\label{tab:prompt_shape_factory}
\end{table*}

\subsection{DayTrader}

Table~\ref{tab:prompt_daytrader} presents the task prompt used in the DayTrader experiment.

\begin{table*}[t]
\centering
\begin{booktabs}{
    width=\textwidth,
    colspec={X[l]},
    cell{1}{1}={c},
}
\toprule
{\textbf{Prompt for DayTrader}}\\
\midrule
\midrule

\textbf{Experiment Rules}\\
- Participate in a repeated investment game called \textit{DayTrader}.\\
- Each participant starts with a fixed amount of money and makes investment decisions across 30 rounds.\\
- Discussion phases occur after rounds 5, 10, 15, 20, 25, and 30; all other rounds proceed directly to the next decision round.\\
- Investments can be made individually or collectively through a group pool.\\
- Individual investments return double the invested amount and benefit only the investor.\\
- Group investments are pooled with contributions from all participants choosing the group option. The pooled amount is tripled and then divided equally among all participants.\\
- Beginning in round 2, the participant(s) with the highest earnings in a round receive a \$90 bonus, split equally in the event of ties.\\

\\

\textbf{Experiment Goal}\\
- Maximize personal monetary balance through strategic investment decisions.\\

\\

\textbf{Available Actions}\\
- \texttt{message}: communicate with other participants.\\
- \texttt{make\_individual\_investment}: invest independently.\\
- \texttt{make\_group\_investment}: invest in the shared group pool.\\

\\

\textbf{Planning Instructions}\\
- Base decisions on persona traits, prior interactions, observed behaviors, and current game status.\\
- During each decision phase, choose at most one investment action (individual or group).\\
- Investment amounts must remain within the configured bounds and available funds.\\
- Discussion phases may be used to coordinate, negotiate, or exchange strategic information.\\
- If no beneficial action is available, return an empty action list.\\
- Adapt strategy after failed or rejected actions and avoid repeating invalid actions.\\

\\

\textbf{Human Behavior Instructions}\\
- Behave like a real participant pursuing personal profit.\\
- Avoid repetitive messages and repeated wording.\\
- Use casual, conversational language rather than formal language.\\
- Maintain continuity across interactions and respond naturally to ongoing discussions.\\
- Consider recent messages and previous agreements when communicating with others.\\

\\

\textbf{Output Requirement}\\
Generate valid action(s) following the predefined JSON action schema.

\\

\bottomrule
\end{booktabs}
\caption{Prompt template used for DayTrader agents. Variables and runtime states are instantiated dynamically during gameplay.}
\label{tab:prompt_daytrader}
\end{table*}

\subsection{Hidden Profiles}

Table~\ref{tab:prompt_hidden_profile} presents the task prompt used in the Hidden Profiles experiment.

\begin{table*}[t]
\centering
\begin{booktabs}{
    width=\textwidth,
    colspec={X[l]},
    cell{1}{1}={c},
}
\toprule
{\textbf{Prompt for Hidden Profile}}\\
\midrule
\midrule

\textbf{Experiment Rules}\\
- Participate in a Hidden Profile group decision-making task.\\
- Each participant receives a partial candidate document containing only a subset of the available information.\\
- Other participants may possess information that is unavailable to you.\\
- The objective is to identify the most suitable candidate from the candidate pool through information sharing and discussion.\\
- Session phases follow the simulation step index: step 1 corresponds to the initial vote, intermediate steps correspond to group discussion, and the final step corresponds to the final vote.\\
- Participants must submit an independent vote both before and after the discussion period.\\

\\

\textbf{Experiment Goal}\\
- Select the most qualified candidate based on all available evidence.\\

\\

\textbf{Experiment Setup and Assignments}\\
- Communication Level: group chat\\
- Candidate Document: \{assigned\_doc\}\\
- Candidate List: \{candidate\_list\}\\
- Participant List: \{participants\_list\}\\

\\

\textbf{Available Actions}\\
- \texttt{message}: contribute to the group discussion.\\
- \texttt{decide}: vote for a candidate (valid only during the initial and final voting phases).\\

\\

\textbf{Planning Instructions}\\
- Base decisions on both your assigned information and evidence revealed by other participants.\\
- Compare candidates according to their strengths, weaknesses, qualifications, and suitability for the role.\\
- Use discussion to exchange information, resolve inconsistencies, and identify missing evidence.\\
- Do not cast votes during the discussion phase.\\
- Submit voting decisions independently rather than following group consensus blindly.\\

\\

\textbf{Human Behavior Instructions}\\
- Behave like a real participant engaged in collaborative decision making.\\
- Maintain continuity with previous discussions and respond naturally to others' messages.\\
- Avoid repetitive messages and repeated arguments.\\
- Use casual, conversational language rather than formal language.\\
- Do not reveal your voting intentions or current voting preferences during discussion.\\
- Participate when your information or perspective is useful, rather than responding to every message.\\
- Focus on sharing unique evidence and avoid repeating information already discussed.\\
- Evaluate candidates critically and avoid expressing excessive agreement without justification.\\

\\

\textbf{Output Requirement}\\
Generate a valid action according to the current session phase.

\\

\bottomrule
\end{booktabs}
\caption{Prompt template used for Hidden Profile agents. Variables enclosed in braces are instantiated dynamically during the experiment.}
\label{tab:prompt_hidden_profile}
\end{table*}

\subsection{The Map Task}

Tables~\ref{tab:prompt_maptask_guide} and~\ref{tab:prompt_maptask_follower}
present the task prompts used for the guide and follower roles in the Map Task experiment.

\begin{table*}[t]
\centering
\begin{booktabs}{
    width=\textwidth,
    colspec={X[l]},
    cell{1}{1}={c},
}
\toprule
{\textbf{Prompt for Map Task Guide Agent}}\\
\midrule
\midrule

\textbf{Experiment Rules}\\
- Participate in a collaborative navigation task called \textit{Map Task}.\\
- Two roles exist: a guide and a follower.\\
- The guide can observe the complete map, including landmarks and the target route.\\
- The follower can observe the landmarks but cannot see the target route.\\
- The two participants must communicate to reproduce the guide’s route on the follower’s map.\\
- Participants cannot directly observe each other’s maps.\\
- In this task, you are assigned the role of \textbf{Guide}.\\

\\

\textbf{Experiment Goal}\\
- Help the follower accurately reconstruct the target route.\\

\\

\textbf{Map Information}\\
- The environment is represented as a discrete grid map.\\
- Some landmarks correspond to blocked regions that cannot be crossed.\\
- Landmark locations and spatial relationships should be used to describe navigation instructions.\\
- The current map is provided as \{\%CURRENT\_MAP\%\}.\\

\\

\textbf{Available Actions}\\
- \texttt{message}: send navigation instructions to the follower.\\

\\

\textbf{Planning Instructions}\\
- Use available landmark information to describe the target route.\\
- Guide the follower step-by-step toward reproducing the route.\\
- Avoid routes that pass through blocked regions.\\
- Adapt instructions based on previous communication and observed progress.\\
- Focus on providing actionable navigation information rather than unrelated conversation.\\

\\

\textbf{Human Behavior Instructions}\\
- Behave like a real participant engaged in collaborative navigation.\\
- Keep messages concise, practical, and easy to follow.\\
- Avoid repetitive instructions and repeated wording.\\
- Do not directly reference grid coordinates, cell indices, or axes.\\
- Use landmarks and relative spatial descriptions instead.\\
- If no new information is available, prefer inaction rather than repeating previous instructions.\\

\\

\textbf{Output Requirement}\\
Generate a valid message that helps the follower reconstruct the route.

\\

\bottomrule
\end{booktabs}
\caption{Prompt template used for Map Task guide agents. Variables enclosed in braces are instantiated dynamically during the experiment.}
\label{tab:prompt_maptask_guide}
\end{table*}

\begin{table*}[t]
\centering
\begin{booktabs}{
    width=\textwidth,
    colspec={X[l]},
    cell{1}{1}={c},
}
\toprule
{\textbf{Prompt for Map Task Follower Agent}}\\
\midrule
\midrule

\textbf{Experiment Rules}\\
- Participate in a collaborative navigation task called \textit{Map Task}.\\
- Two roles exist: a guide and a follower.\\
- The guide can observe the complete map, including landmarks and the target route.\\
- The follower can observe the landmarks but cannot see the target route.\\
- Participants must communicate to reconstruct the guide's route.\\
- Participants cannot directly observe each other's maps.\\
- In this task, you are assigned the role of \textbf{Follower}.\\

\\

\textbf{Experiment Goal}\\
- Reproduce the guide's route as accurately as possible.\\

\\

\textbf{Map Information}\\
- The environment is represented as a discrete grid map.\\
- Blocked landmarks correspond to impassable cells that must never appear in the route.\\
- Landmark positions and spatial relationships should be used to interpret navigation instructions.\\
- Route segments must be 4-connected (up, down, left, right only).\\
- Diagonal movements are invalid and will be rejected.\\
- When diagonal geometry is implied, construct an orthogonal stair-step path instead.\\
- The current map is provided as \{\%CURRENT\_MAP\%\}.\\

\\

\textbf{Available Actions}\\
- \texttt{message}: communicate with the guide.\\
- \texttt{draw}: add route segments to the map.\\
- \texttt{erase}: remove route segments.\\
- \texttt{undo}: revert the most recent route edit.\\
- \texttt{reset}: clear the entire route.\\

\\

\textbf{Planning Instructions}\\
- Follow the guide's instructions while maintaining route validity.\\
- Check the current route state and recent action feedback before drawing.\\
- Avoid blocked cells and invalid connectivity patterns.\\
- Do not draw speculative long routes when instructions are ambiguous.\\
- If a drawing action is rejected, revise the route based on the error feedback rather than repeating the same action.\\
- Use communication to clarify uncertain instructions and confirm progress.\\

\\

\textbf{Human Behavior Instructions}\\
- Behave like a real participant engaged in collaborative navigation.\\
- Keep communication short, practical, and conversational.\\
- Avoid repetitive messages and repeated route edits.\\
- Show reasonable uncertainty when instructions are unclear.\\
- Proactively suggest likely next steps instead of remaining silent.\\
- Briefly acknowledge mistakes and continue solving the task.\\
- Avoid emoji and overly formal language.\\

\\

\textbf{Output Requirement}\\
Generate a valid action according to the current map state and interaction history.

\\

\bottomrule
\end{booktabs}
\caption{Prompt template used for Map Task follower agents. Variables enclosed in braces are instantiated dynamically during the experiment.}
\label{tab:prompt_maptask_follower}
\end{table*}

\end{document}